\documentclass[lettersize,journal]{IEEEtran}
\usepackage{amsmath,amsfonts}
\usepackage{algorithmic}
\usepackage{algorithm}
\usepackage{array}
\usepackage[caption=false,font=normalsize,labelfont=sf,textfont=sf]{subfig}
\usepackage{textcomp}
\usepackage{stfloats}
\usepackage{url}
\usepackage{verbatim}
\usepackage{graphicx}
\usepackage{cite}
\usepackage{bbding}
\usepackage{hyperref}
\usepackage{amssymb}

\newtheorem{definition}{Definition}
\newtheorem{remark}{Remark}
\usepackage{multirow}
\usepackage{booktabs}

\begin{document}

\title{All-around Neural Collapse for Imbalanced Classification}

\author{Enhao~Zhang,
        Chaohua~Li,
        Chuanxing~Geng,    
        and~Songcan~Chen,~\IEEEmembership{Senior~Member,~IEEE}% <-this % stops a space
\IEEEcompsocitemizethanks{\IEEEcompsocthanksitem E. Zhang, C. Geng, C. Li, and S. Chen are with MIIT Key Laboratory of Pattern Analysis and Machine Intelligence, China
and College of Computer Science and Technology, Nanjing University of Aeronautics and Astronautics (NUAA), Nanjing 211106, China\protect\\
E-mail:\{zhangeh, chaohuali, gengchuanxing, s.chen\}@nuaa.edu.cn.}

\thanks{This paper was produced by the IEEE Publication Technology Group. They are in Piscataway, NJ.}% <-this % stops a space
\thanks{Manuscript received April 19, 2021; revised August 16, 2021.}}

\markboth{Journal of \LaTeX\ Class Files,~Vol.~14, No.~8, August~2021}%
{Shell \MakeLowercase{\textit{et al.}}: A Sample Article Using IEEEtran.cls for IEEE Journals}


\maketitle

\begin{abstract}
Neural Collapse (NC) presents an elegant geometric structure that enables individual activations (features), class means and classifier (weights) vectors to reach \textit{optimal} inter-class separability during the terminal phase of training on a \textit{balanced} dataset. Once shifted to imbalanced classification, such an optimal structure of NC can be readily destroyed by the notorious \textit{minority collapse}, where the classifier vectors corresponding to the minority classes are squeezed. In response, existing works endeavor to recover NC typically by optimizing classifiers. However, we discover that this squeezing phenomenon is not only confined to classifier vectors but also occurs with class means.
    Consequently, reconstructing NC solely at the classifier aspect may be futile, as the feature means remain compressed, leading to the violation of inherent \textit{self-duality} in NC (\textit{i.e.}, class means and classifier vectors converge mutually) and incidentally, resulting in an unsatisfactory collapse of individual activations towards the corresponding class means. To shake off these dilemmas, we present a unified \textbf{All}-around \textbf{N}eural \textbf{C}ollapse framework (AllNC), aiming to comprehensively restore NC across multiple aspects including individual activations, class means and classifier vectors. We thoroughly analyze its effectiveness and verify on multiple benchmark datasets that it achieves state-of-the-art in both balanced and imbalanced settings.
\end{abstract}


\section{Introduction}
\label{sec:intro}
\IEEEPARstart{D}{eep} neural networks (DNNs) have achieved remarkable success on various computer vision tasks~\cite{he2016deep}. However, the success of such over-parameterized networks still lacks a comprehensive understanding~\cite{jiang2023generalized}. In a recent study, Papyan et al.~\cite{papyan2020prevalence} empirically observe that, during the terminal phase of training on \textit{balanced} datasets, DNNs tend to form a specific geometrical structure, formalized as \textit{Neural Collapse} (NC). When the model reaches zero training error, NC unveils a highly symmetric geometric structure arising from the last layer activation map.  Specifically, i) individual activations collapse towards their respective class means, ii) all class means are aligned to the vertices of a simplex \textit{Equirectangular Tight Frame} (ETF), and iii) the classifier vectors and class means converge towards each other, satisfying so-called \textit{self-duality} property. This convergence yields an optimal linearly separable representation for the classification task. Subsequent studies reveal this phenomenon's physical mechanisms, deducing that the NC solution is globally optimal under certain constraints~\cite{tirer2022extended}. 
\begin{figure}[!t]
  \centering
   \includegraphics[width=0.75\linewidth]{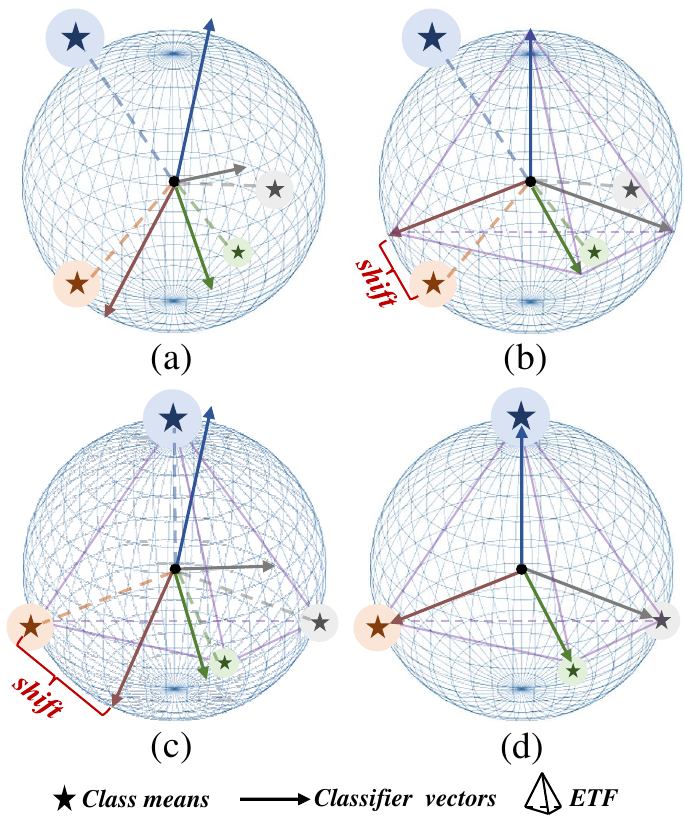}
   \caption{Geometric illustration of class means and classifier vectors. (a) Plain cross-entropy loss. (b) ETF classifier. (c) Embedded feature constraints. (d) Our method. In particular, different colors indicate various classes, where the size of the class mean legend is proportional to the corresponding intra-class sample number.}
   \label{fig1}
\end{figure}

Unfortunately, when being up against an imbalanced dataset, typically such as long-tailed distribution data, where a few majority classes occupy most of the samples~\cite{cao2021learning,liang2023tackling,zhang2023deep}, one often suffers from the annoying \textit{minority collapse}~\cite{fang2021exploring}. This refers to the squeezing of classifier vectors belonging to minority classes, which erodes the NC phenomenon and fundamentally limits the performance of deep learning, especially in the minority classes~\cite{thrampoulidis2022imbalance}. Further, Liu et al. elaborate that the NC phenomenon can assist in minimizing the generalization error in the long-tailed problem~\cite{liu2023inducing}. Therefore, recent researches endeavor to restore the NC phenomenon by enhancing classifier~\cite{kasarla2022maximum,yang2022inducing}. 
Nevertheless, such one-sided boosting is inadequate since we discover that besides the classifier vectors, the class means of the minority classes are likewise squeezed, coined as \textit{complete minority collapses}, as shown in Fig. \ref{fig1} (a).
Although improving the classifier can encourage its components to align with the ideal ETF structure, the corresponding class means remain corrupted, leading to a significant gap between them and the classifier, as depicted in Fig.\ref{fig1}(b). As well, by imposing constraints solely on the latent representations, the learned model is still compromised, as evidenced in Fig.\ref{fig1}(c), where the classifier vectors of the minority classes remain compressed.
Consequently, reconstructing the NC at any single aspect alone may cause the trained models to lose sight of the other, destroying the inherent \textit{self-duality} property of NC, which inevitably brings about a sharp shift in the embedded features and classifier vectors during the projection process.

To address the aforementioned issues, this paper introduces a comprehensive end-to-end solution named \textbf{All}-around \textbf{N}eural \textbf{C}ollapse (AllNC) to recover the NC phenomenon in an imbalanced context from multiple aspects, including individual activations, class means and classifier vectors. 
Specifically, at the individual activation aspect, while alignment with the class mean can occur naturally, regardless of data distribution, achieving satisfactory aggregation results is challenging in practice~\cite {liu2023inducing}.
For this reason, existing studies employ supervised contrastive learning for imbalanced classification tasks, yet they are vulnerable to encountering  \textit{dual class-imbalance issue}, \textit{i.e.}, the degree of imbalance becomes quadratic~\cite{zhong2022rebalanced}. As a solution, we propose a Hybrid Contrastive loss (HyCon) without negative samples, which requires only two views of the anchor point and the corresponding class mean to prompt individual activations to obtain the \textit{variability collapse} (\textit{i.e.}, the intra-class covariance tends to zero) in NC. Furthermore, at both the class means and the classifier vectors aspects, to fulfill the \textit{self-duality} property, we introduce a straightforward yet effective Peer-to-Peer loss (P2P) that simultaneously aligns the class means and the classifier vectors to a simplex ETF, thereby achieving optimal inter-class separability. Finally, to reconcile the learning of individual activations and classifiers, we propose a unified Generalized Bilateral-Branch Network (GBBN) that progressively decouples model training over time without a cumbersome two-stage scheme.

Extensive experiments demonstrate that with the three complementary components of HyCon, P2P and GBBN, AllNC not only recovers the desired NC phenomenon in imbalanced training (illustrated in Fig. \ref{fig1} (d)) but also surpasses other approaches by a large margin. The main contributions of our work can be summarised as follows: 

\begin{itemize}
    \item To the best of our knowledge, we are the first to discover that training imbalanced data encounters \textit{complete minority collapses}, and observe that recovering NC at only one aspect cannot realize inherent \textit{self-duality} property.
    \item We propose an end-to-end learning program, AllNC, which facilitates the learned individual activations, class means and classifier vectors to cater for the NC phenomenon simultaneously, and improves model generalization via progressively decoupled feature and classifier training on a single branch.
    \item Our method achieves state-of-the-art performance on multiple imbalanced classification benchmark datasets. In addition, we also validate the effectiveness of AllNC in a balanced setup.
\end{itemize}

\section{Related Works}
\label{sec:relate}

\subsection{Imbalanced Classification}

Conventional imbalanced classification methods primarily focus on designing class re-balancing strategies via compromising between the number of samples (features) in the head and tail classes, \textit{e.g.}, re-sampling~\cite{ando2017deep,shen2016relay,buda2018systematic,xie2020gaussian,chen2023generating}, or designing losses that match the test distribution, e.g., re-weighting~\cite{cao2019learning,cui2019class,li2022equalized,li2024multi}. However, these methods may result in sub-optimal solutions~\cite{cui2019class}. Subsequent studies simulated a balanced representation space by improving the SupCon loss \cite{khosla2020supervised} to obtain consistency between pairs of positive samples~\cite{zhu2022balanced,li2022targeted,cui2021parametric,10255367}. Nevertheless, \cite{zhong2022rebalanced} pointed out that, burdened by a large pool of negative samples, the supervised contrastive framework suffers from a serious \textit{dual class-imbalance issue}.
For this reason, we propose HyCon without negative samples, which not only inherits the consistency between pairs of positive samples but also aligns anchors with the corresponding class mean.

In addition, recent studies have found that decoupling feature learning and classifier training using a two-stage approach yields superior features\cite{cao2019learning,kang2019decoupling}. For example, \cite{zhou2020bbn} introduces a bilateral-branch network to adjust the representation learning and the classifier through a cumulative learning strategy. Inspired by this, we propose the Generalized Bilateral-Branch Network (GBBN) with \textit{only one branch} for cumulative learning and regulating the learning tendency according to the degree of imbalance. 

\subsection{Neural Collapse}
NC can provide an elegant geometric structure simultaneously for individual activations, class means and classifier vectors in the terminal phase of training, allowing them to attain a state of optimal separability among classes~\cite{papyan2020prevalence}. Recent studies have demonstrated that NC to some extent also occurs in most of the intermediate hidden layers of the network\cite{parker2023neural,rangamani2023feature}. Although numerous studies have both empirically and theoretically validated the effectiveness of NC, once shifted to imbalanced scenarios, the classifier vectors of the minority classes are squeezed, inevitably leading to the destruction of this optimal structure \cite{fang2021exploring}.
Therefore, existing imbalance methods wish to reproduce the NC phenomenon by improving the classifier to align it with the ETF structure \cite{saeed2024hyperspherical, seo2024learning}. For instance, \cite{yang2022inducing} employed a fixed ETF classifier instead of a learnable classifier, attempting to alleviate the \textit{minority collapse}.
 However, with our latest observation of \textit {complete minority collapses}, it is obvious that these one-sided efforts are far inadequate. Distinct from existing works, this paper proposes a comprehensive solution to simultaneously address the challenges posed by imbalanced data in both the feature space and the classifier, allowing NC to occur naturally, irrespective of whether the dataset is balanced or not.

 \section{Problem Setup}
In this section, we provide more details about the NC and present our new findings about the dilemmas faced by NC in an imbalanced context.

\subsection{Preliminaries}
Papyan et al.~\cite{papyan2020prevalence} revealed the NC phenomenon, whereby in the terminal phase of training on a balanced dataset, the individual activations of the last layer converge to the corresponding intraclass mean, and the within-class means together with the classifier vectors collapse to the vertices of a  simplex ETF. First, we describe the definition of a simplex ETF on simplex as follows

\begin{definition}[\textbf{Simplex ETF}]
\label{ETF}
    A collection of vectors $\boldsymbol{v_c} \in \mathbb{R}^q, c=1,2, \cdots, C$ is said to be a simplex equiangular tight frame if: 
\begin{eqnarray}
 \boldsymbol{V}=\sqrt{ \frac{C}{C-1} } \boldsymbol{U} \left( \boldsymbol{I_C} - \frac{1}{C} \boldsymbol{1_C 1^T_C}  \right) ,
\end{eqnarray}
where $\boldsymbol{V}=\left[\boldsymbol{v_1}, \cdots, \boldsymbol{v_C}\right] \in \mathbb{R}^{q \times C}, \boldsymbol{U} \in \mathbb{R}^{q \times C}$ allows a rotation and satisfies $\boldsymbol{U^TU=I_C}$,  $\boldsymbol{I_C}$ is the identity matrix and $\boldsymbol{1_C}$ is an all-ones vector.
\end{definition}

Consequently, the simplex ETF obeys for $\forall i,j\in [1, C]$,

\begin{eqnarray}
\label{2}
 \begin{aligned}
      \boldsymbol{v_{i}}^{T}\boldsymbol{{v}_{j}}=\left\{ \begin{matrix}
   \text{      }1,\text{    }&i=j  \\
   -\frac{1}{C-1}, &i\ne j \end{matrix} \right..
 \end{aligned}
 \label{etf}
\end{eqnarray}
Then, the NC phenomenon can be characterized by four manifestations in the classifier and last-layer activations~\cite{papyan2020prevalence}:  
\begin{figure*}
  \centering
   \includegraphics[width=0.95\linewidth]{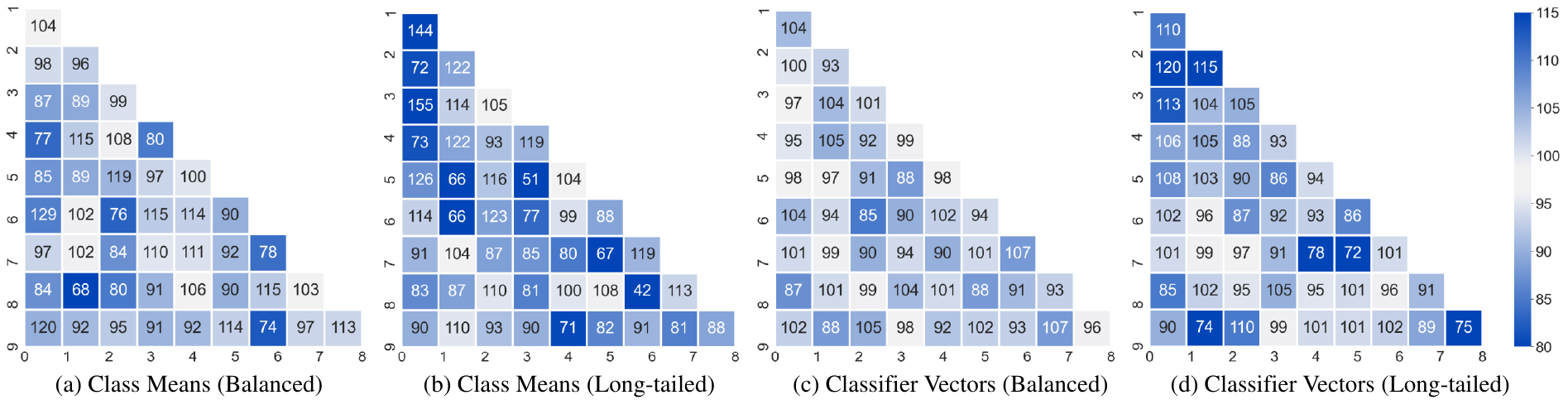}
   \caption{Inter-Class Pairwise Angles (ICPAs) corresponding to the centered class means and classifier vectors in CIFAR10 and CIFAR10-LT, where a grid represents an interclass angle, with the darker-blue grid means a more serious deviation from the optimal ICPA(\textit{i.e.}, $96.4^\circ$).}
   \label{nc2_fig}
\end{figure*}

\begin{itemize}
    \item {\textbf{(NC1) Variability collapse:} the intra-class variance collapses to zero as these activations collapse to their class means: 
    \begin{eqnarray}
    \boldsymbol{\Sigma_{W}} \to 0,
    \end{eqnarray}
    where $\boldsymbol{\Sigma}_W \triangleq \underset{i, c}{\operatorname{Ave}}\left\{\left(\boldsymbol{z}_{i, c}-\boldsymbol{\mu}_c\right)\left(\boldsymbol{z}_{i, c}-\boldsymbol{\mu}_c\right)^{\top}\right\}$, $\boldsymbol{\mu}_c \triangleq \underset{i}{\operatorname{Ave}}\left\{\boldsymbol{z}_{i, c}\right\}, \quad c=1, \ldots, C$, is the mean of the $c$-th class.  $\boldsymbol{z}_{i, c}$ is the last layer of activations of the $i$-th sample in the $c$-th class.}
    \item {\textbf{(NC2) Convergence to simplex ETF:} the class centers centered by the global mean will converge to the vertices of a simplex ETF as defined in Definition \ref{ETF}, \textit{i.e.}, $\left \{ \tilde{\boldsymbol{\mu}_{c}} \right \} _{c=1}^{C}$ satisfy Eq. \ref{2},
    where $\tilde{\boldsymbol{\mu}}_c \triangleq \left(\boldsymbol{\mu}_c-\boldsymbol{\mu}_G\right) /\left\|\boldsymbol{\mu}_c-\boldsymbol{\mu}_G\right\|_2$, $\boldsymbol{\mu}_G \triangleq \underset{i,c}{\operatorname{Ave}}\left\{\boldsymbol{z}_{i, c}\right\}$.}
    \item {\textbf{(NC3) Convergence to self-duality:} Class means and classifier vectors converge to each other, up to rescaling: 
    \begin{eqnarray}
\Delta =\left\|\frac{\boldsymbol{W}^{\top}}{\|\boldsymbol{W}\|_F}-\frac{\dot{\boldsymbol{M}}}{\|\dot{\boldsymbol{M}}\|_F}\right\|_F \rightarrow 0.
\label{nc3}
    \end{eqnarray}
    where $\dot{\boldsymbol{M}}=\left[\boldsymbol{\mu}_c-\boldsymbol{\mu}_G, c=1, \ldots, C\right]$, $\boldsymbol{W}=\left[\boldsymbol{\omega}_1, \ldots, \boldsymbol{\omega}_C\right]$ denotes the classifier vector matrix. \textit{This ensures minimal confusion between two classes}. 
    }
        \item {\textbf{(NC4) Simple decision rule:} When (NC1)-(NC3) hold, the model prediction can be simplified to the nearest class center prediction:
    \begin{eqnarray}
\arg \max _{c^{\prime}}\left\langle\boldsymbol{\omega}_{c^{\prime}}, \boldsymbol{h}\right\rangle+b_{c^{\prime}} \rightarrow {\arg \min_{c^{\prime}}}\left\|\boldsymbol{h}-\boldsymbol{\mu}_{c^{\prime}}\right\|_2,
\label{nc4}
    \end{eqnarray}
    where $b$ denotes the bias term and $\boldsymbol{h}$ is the last-layer feature of a sample for prediction.}
    
\end{itemize}

\subsection{Imbalance Dilemmas}
\label{ID}
In this part, we first demonstrate the new finding of \textit{complete minority collapses}. Then, we analyze the impact of recovering the NC phenomenon only at the classifier aspect on the intrinsic \textit {self-duality} property.

\subsubsection{Complete Minority Collapses}
\label{CMC}
In light of NC2 and NC3, it can easily be derived via Eq.~\ref{2} that the optimal \textbf{I}nter-\textbf{C}lass \textbf{P}airwise \textbf{A}ngle (ICPA) related to the learned class means and classifier vectors both equals $-\frac{1}{C-1}$. Next, we compute ICPAs for the class means (after zero-centered normalization) and classifier vectors specific to CIFAR10 and CIFAR10-LT (\textit{i.e.}, long-tailed version, where the sample size within the class is exponentially decreasing within class imbalance rate $\beta=100$.), respectively. The ICPAs' calculation results are displayed in Fig. \ref{nc2_fig}, where the darker-blue grid means a more serious deviation from the optimal ICPA, which is equal to  $-\frac{1}{10-1} \approx 96.4^\circ$.

It is evident from Fig. \ref{nc2_fig} that the lighter colors of the grids in Figs. (a) and (c) indicate that the ICPAs corresponding to the class means and classifier vectors obtained from the balanced dataset are close to the optimal ICPA, in line with the descriptions in NC. 
Conversely, most grids in Figs. (b) and (d) are darker, suggesting a significant gap between the ICPAs computed in the imbalanced context and the optimal ICPA. 
And the majority classes typically possess larger ICPAs, while the reverse is true for the minority classes. 
This observation reveals that \textit{when faced with class imbalance, both the class means and classifier vectors of the minority classes are subject to compression by the majority classes}, which is identified as \textit{Complete Minority Collapses}. 

\begin{figure*}[h]
  \centering
   \includegraphics[width=0.95\linewidth]{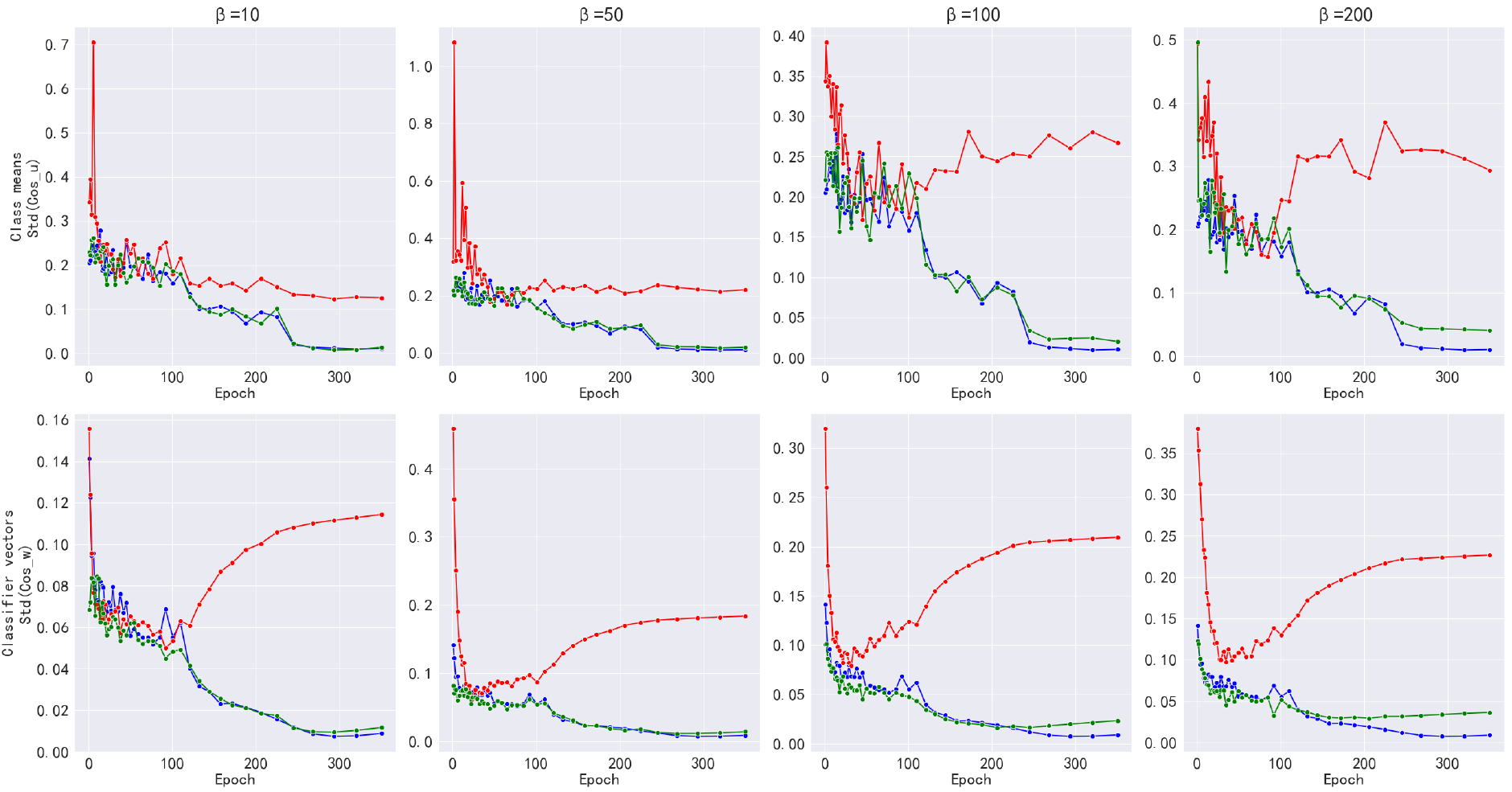}
   \caption{The Std of the cosines between all pairs of different classes of the class mean/classifier vector on CIFAR10/CIFAR10-LT. The blue lines represent the Std obtained using CE loss on balanced data, while the red and green lines represent the Std calculated using the CE loss and our proposed AllNC on a long-tailed dataset with an imbalance rate of $\beta$, respectively.}
   \label{avg_cos}
\end{figure*}

To further validate the observed \textit{complete minority collapses} phenomenon, we use ResNet18 as a backbone to compute the standard deviation (Std) of the cosine between pairs of centered class means and classifier vectors across all distinct pairs of classes $c$ and $c^{\prime}$ on CIFAR10/CIFAR10-LT.
Mathematically, this can be expressed as:

\begin{eqnarray}
\label{cos_u}
    \cos _{\boldsymbol{\mu}}\left(c, c^{\prime}\right)=\frac{\left\langle\boldsymbol{\mu}_c-\boldsymbol{\mu}_G, \boldsymbol{\mu}_{c^{\prime}}-\boldsymbol{\mu}_G\right\rangle}{\left(\left\|\boldsymbol{\mu}_c-\boldsymbol{\mu}_G\right\|_2\left\|\boldsymbol{\mu}_{c^{\prime}}-\boldsymbol{\mu}_G\right\|_2\right)},
\end{eqnarray}

\begin{eqnarray}
\label{cos_w}
    \cos _{\boldsymbol{\omega}}\left(c, c^{\prime}\right)=\frac{\left\langle\boldsymbol{\omega}_c-\boldsymbol{\omega}_G, \boldsymbol{\omega}_{c^{\prime}}-\boldsymbol{\omega}_G\right\rangle}{\left(\left\|\boldsymbol{\omega}_c-\boldsymbol{\omega}_G\right\|_2\left\|\boldsymbol{\omega}_{c^{\prime}}-\boldsymbol{\omega}_G\right\|_2\right)}
\end{eqnarray}
    where $c, c^{\prime}=1, \ldots, C$ and $c \ne  c^{\prime}$. It is not difficult to find that when the class means/classifier vectors have the same interclass pairwise angles, then the corresponding cosine values are approximately equal, such that the obtained Std tends to 0, and vice versa.

    As indicated in Fig. \ref{avg_cos}, we observe that the Stds corresponding to the class means (top) and classifier vectors (bottom) obtained using the balanced dataset converge to 0 at the terminal stage of training (blue lines). This indicates that all the ICPAs of the class means and classifier vectors are approximately equal, implying that they align with the ideal simplex ETF structure.
     However, in the imbalance context, we can discover that the Std obtained for both the class means and the classifier vectors are much larger than 0 (red lines), and this tendency gradually increases as the degree of imbalance and epoch increase. 
     This is because the trained model's emphasis on the majority classes causes an increase in their corresponding ICPAs, while the ones corresponding to the minority classes decrease. It is further shown that\textit{ during imbalanced training, both the class means and classifier vectors corresponding to the minority classes are squeezed, leading to complete minority collapses.} Furthermore, this phenomenon becomes more severe as the degree of imbalance and the number of training sessions increase.

\subsubsection{Unsatisfied Self-Duality}
\label{USD}
To remediate the issue of \textit{minority collapse}, contemporary research typically opts to upgrade classifiers to recover NC, yet these endeavors might be futile. The occurrence of \textit{complete minority collapses} hinders the realization of the intrinsic \textit{self-duality} of NC. We will verify this phenomenon through a TOY experiment in the subsequent discussion.

\begin{figure}
  \centering
   \includegraphics[width=0.95\linewidth]{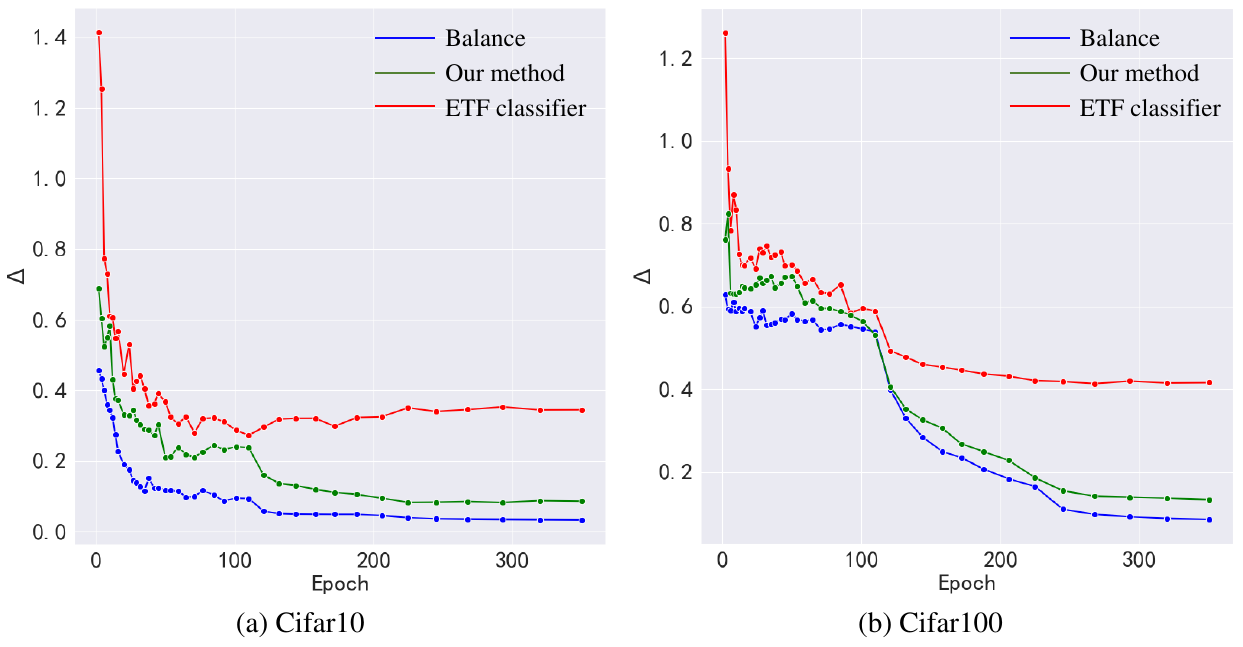}
   \caption{Self-Duality Metrics obtained on CIFAR10/100 and CIFAR10/100-LT, where the red and green lines indicate the results obtained in the long-tailed datasets.}
   \label{nc3_fig}
\end{figure}

Following NC3, we use Eq. \ref{nc3} as a \textit{Self-Duality Metrics}, which when $\Delta \rightarrow 0$ indicates that both the class means and the classifier vectors converge to each other, satisfying the \textit{self-duality} property. Next, we respectively train the models using CIFAR10/100 and CIFAR10/100-LT and compute the corresponding self-duality metrics. 
In particular, we apply cross-entropy loss and generic classifiers in the \textit{balanced} dataset as well as point regression loss and an ETF classifier (attempting to recover NC solely at the classifier aspect.)\cite{yang2022inducing} in the \textit{long-tailed} dataset.
From Fig. \ref{nc3_fig}, we can observe intuitively that self-duality metrics obtained from a balanced dataset tend towards zero, whereas those obtained from an ETF classifier on a long-tailed dataset are significantly higher. This indicates that \textit{the recovery of NC at the classifier aspect alone fails to meet the self-duality property}. The reason is that although classifier vectors align with the ideal ETF structure, the corresponding feature means are still corrupted and there is a significant gap between them, which fundamentally limits the model's performance.

\begin{figure*}[!t]
  \centering
   \includegraphics[width=0.95\linewidth]{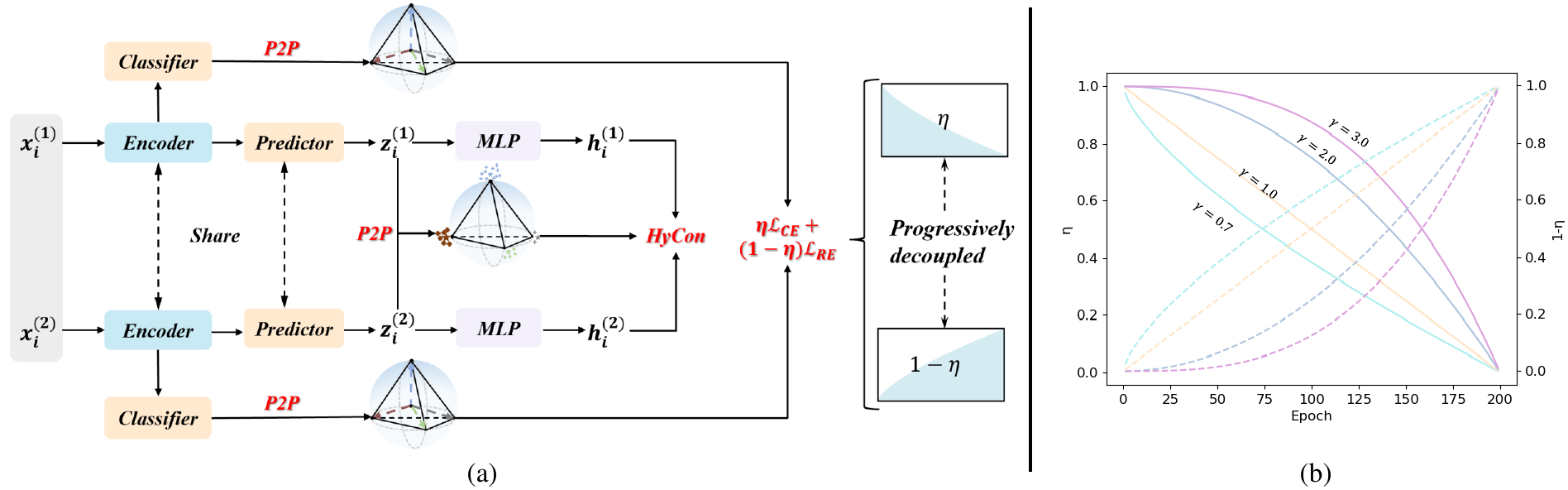}
   \caption{(a) The architecture diagram of AllNC, which utilizes a contrastive framework and two additional classification heads. In this design, the encoders, predictors, MLPs and classifiers between the two branches share the same weights, largely reducing the computational complexity during the inference phase. (b) Trend plots of parameters $\eta$ (solid lines) and $1-\eta$ (dashed lines) for the adjustable parameter $\gamma$.}
   \label{framework}
\end{figure*}

\section{All-around Neural Collapse}

In this section, we delve into the details of our work. In Sec. \ref{HyCon}, we propose the HyCon,  which aims to ensure individual activations satisfy NC1. In Sec. \ref{Peer-to-Peer}, we introduce the P2P, which aligns both the class means and the classifier vectors with the ideal ETF structure, thereby fulfilling both NC2 and NC3. Finally, in Sec. \ref{GBBN}, we design a unified GBBN framework to progressively decouple the learning of both the features and the classifiers. The general flowchart of our approach is illustrated in Fig. \ref{framework}(a).

\subsection{Hybrid Contrastive Loss}
\label{HyCon}
To obtain clusters with both inter-class separation and intra-class compactness, existing methods commonly introduce contrastive learning to constrain the feature space. Nevertheless, they typically require a large number of negative samples, leading to the \textit{dual class-imbalance issue}. 
In response to this problem, inspired by SimSiam~\cite{chen2021exploring}, we propose HyCon without requiring negative samples, which achieves contrastive consistency by maximizing the cosine similarity between the anchor point and the corresponding augmented positive sample. Different from SimSiam, HyCon additionally introduces labeling information to guide the anchor point to align to the corresponding class mean, enabling the within-class variance to converge to zero.

Specifically, two views $\boldsymbol{x}^{(1)}$ and $\boldsymbol{x}^{(2)}$ of the same anchor image $\boldsymbol{x}$ are passed through the encoder and the projection head to obtain their representations $\boldsymbol{z}^{(1)}$ and $\boldsymbol{z}^{(2)}$, with corresponding intra-class means $\boldsymbol{u}^{(1)}$ and $\boldsymbol{u}^{(2)}$. Then the obtained representations are transformed into the outputs $\boldsymbol{h}^{(1)}$ and $\boldsymbol{h}^{(2)}$ after MLP. HyCon can be formalized as:
\begin{equation}
\mathcal{L}_{con}=sim(\boldsymbol{h}^{(1)},\boldsymbol{u}^{(2)},sg(\boldsymbol{z}^{(2)}))+sim(\boldsymbol{h}^{(2)},\boldsymbol{u}^{(1)},sg(\boldsymbol{z}^{(1)})),
\label{hycon_loss}
\end{equation}%
where
\begin{equation}
    sim(\boldsymbol{h},\boldsymbol{u},sg(\boldsymbol{z}))=-\frac{\boldsymbol{h}}{\left || \boldsymbol{h} \right || } \cdot \frac{\boldsymbol{z}}{\left || \boldsymbol{z} \right || }  - \frac{\boldsymbol{u}}{\left || \boldsymbol{u} \right || } \cdot \frac{\boldsymbol{z}}{\left || \boldsymbol{z} \right || }, \nonumber
\end{equation}
$\left || \cdot \right ||$ is $l_{2}$ normalization. Following
SimSiam, $sg(\boldsymbol{z})$ indicates the stopping gradient operation to prevent model collapse. Unlike existing long-tailed methods based on supervised contrastive learning, HyCon eliminates the need for additional positive and negative samples, thus markedly reducing the model's complexity. 
\begin{remark} It is easy to identify that the loss $\mathcal{L}_{con}$ is minimized if and only if $cos(\boldsymbol{h},\boldsymbol{z}) = 1$ and $cos(\boldsymbol{u},\boldsymbol{z}) = 1$, which means that individual activations are aligned with their corresponding class means. Therefore, we can obtain  $\boldsymbol{\Sigma}_W \triangleq \underset{i, c}{\operatorname{Ave}}\left\{\left(\boldsymbol{z}_{i, c}-\boldsymbol{\mu}_c\right)\left(\boldsymbol{z}_{i, c}-\boldsymbol{\mu}_c\right)^{\top}\right\} \to 0 $, \textit{i.e.}, NC1 is satisfied. In addition, the visualization results of \ref{reproduction} further validate that our method can reproduce the property.
\end{remark}

\subsection{Peer-to-Peer Loss}
\label{Peer-to-Peer}
As revealed in Sec.\ref{ID}, the existence of the \textit{complete minority collapses} leads to the common approach of solely recovering NC at the classifier aspect, which fails to satisfy the inherent \textit{self-duality} property. Therefore, we propose a simple and effective P2P loss that simultaneously constrains class means and classifier vectors to converge to the ideal simplex ETF.

According to Definition \ref{ETF}, if the learned class means $\left\{\tilde{\boldsymbol{\mu}}_1, \tilde{\boldsymbol{\mu}}_2, \dots,\tilde{\boldsymbol{\mu}}_C \right\}$ and classifier vectors $\left\{{\boldsymbol{\omega}}_1, {\boldsymbol{\omega}}_2, \dots,{\boldsymbol{\omega}}_C \right\}$ satisfy  the ETF structure, their corresponding optimal ICPAs both are
\begin{equation}
\rho _{ij}\triangleq \frac{C}{C-1} \delta_{i, j}-\frac{1}{C-1},
\end{equation}
where $\delta_{i,j}$ is the Kronecker delta symbol. We then align the class mean with the simplex ETF using the P2P loss, which can be formulated as:
\begin{equation}
\label{lpu}
\mathcal{L}_{p}(\tilde{\boldsymbol{\mu}})=\frac{1}{C^2} \sum_{i,j=1}^{C} (\tilde{\boldsymbol{\mu}}_i^T \tilde{\boldsymbol{\mu}}_j-\rho _{ij})^2.
\end{equation}

\begin{remark}
We can easily spot that the loss in Eq. \ref{lpu} reaches optimality ($\mathcal{L}_{p}(\tilde{\boldsymbol{\mu}})=0$) if and only if $\tilde{\boldsymbol{\mu}}_i^T \tilde{\boldsymbol{\mu}}_j = \rho _{ij}$, which implies that the class centers converge to the ideal simplex ETF structure, satisfying the conditions of NC2.
\end{remark}

Furthermore, we extend the P2P loss to the classifier vector by the following formulation Eq.~\ref{lpw},
\begin{equation}
\label{lpw}
\mathcal{L}_{p}(\boldsymbol{\omega})=\frac{1}{C^2} \sum_{i,j=1}^{C}(\boldsymbol{\omega}_i^T \boldsymbol{\omega}_j-\rho _{ij})^2.
\end{equation}

\begin{remark}
When both Eq. \ref{lpu} and \ref{lpw} reach their optimal solutions, it follows that $\tilde{\boldsymbol{\mu}}_i^T \tilde{\boldsymbol{\mu}}_j = \rho _{ij} = \boldsymbol{\omega}_i^T \boldsymbol{\omega}_j$. This implies that both are equivalent in terms of ICPAs and converge towards each other, thus satisfying the \textit{self-duality} property. 
Moreover, as observed in Fig. \ref{nc3_fig}, our method (green lines) significantly decreases the self-duality metrics to close to 0, further affirming that in practical application, the proposed method achieves convergence of the class means with the classifier vectors to \textit{self-duality}.
\end{remark}

\subsection{Generalized Bilateral-Branch Network}
\label{GBBN}
To bypass using a cumbersome two-stage approach to decouple individual activations and classifier training, we propose a unified GBBN learning framework, which progressively decouples individual activations and classifier training over time, gradually switching the training focus to the minority classes. In particular, we define $\mathcal{L}_{ce}$ as the CE loss and $\mathcal{L}_{re}$ as any rebalancing loss. For given augmented inputs $\left\{\left(\boldsymbol{x_i^{(1)}}, \boldsymbol{y_i}\right)\right\}_{i=1}^N$, then a branching loss in the GBBN can be expressed as
\begin{align}
\label{cls}
\mathcal{L}_{cls}^{(1)} = &\frac{1}{N} \sum_{i=1}^{N} \eta \mathcal{L}_{ce}(\boldsymbol{x_i^{(1)}}, \boldsymbol{y_i}) +  (1-\eta) (\mathcal{L}_{re}(\boldsymbol{x_i^{(1)}}, \boldsymbol{y_i}) \nonumber \\
 + &\mathcal{L}_{p}(\boldsymbol{\omega})),
\end{align}%
where $\eta$ is automatically generated based on the training epoch. Specifically, $T_{max}$ denotes the total number of training epochs, and $T$ represents the current iteration count.
 $\eta$ is calculated by
\begin{equation}
\eta = 1-\left (\frac{T}{T_{max}} \right)^ \gamma,
\end{equation}
where $\gamma$ is an adjustable parameter. 
Based on Eq. \ref{cls}, in a branch, GBBN can adaptively achieve progressively decoupled feature learning and classifier learning with respect to the weight $\eta$. 
Furthermore, by modifying the parameter $\gamma$, we can adjust the training favoring of the model. 
As shown in Fig. \ref{framework}(b), it is evident that as $\gamma$ increases, the model emphasizes feature learning more, and vice versa.

Finally, the total loss of our AllNC is defined as,
\begin{equation}
\mathcal{L}_{all} = \mathcal{L}_{cls}^{(1)} + \mathcal{L}_{cls}^{(2)} + \alpha (\mathcal{L}_{con} + \mathcal{L}_{p}(\tilde{\boldsymbol{\mu}})),
\label{all_loss}
\end{equation}
where $\alpha$ is a hyperparameter. The overall training procedure of the method proposed in this paper is summarised in Algorithm~\ref{alg}.

\begin{algorithm}
    \caption{Training procedure of AllNC}
    \label{alg}
    \textbf{Input}: Training data $\mathcal{D}=\left\{\left(\boldsymbol{x}_i, y_i\right)\right\}_{i=1}^N$\\
    \textbf{Parameter}: Feature encoder $\phi (\cdot )$; projection $\textit{proj}_{1}(\cdot)$, $\textit{proj}_{2}(\cdot)$; classifier $\textit{f}(\cdot)$, adjustable parameters $\gamma,\alpha$; 
    
    \begin{algorithmic}[1] %[1] enables line numbers
        \STATE Initialize model parameters $\phi, \textit{proj}_{1}(\cdot), \textit{proj}_{2}(\cdot), \textit{f}$;
        \FOR{$T=1$ in $T_{max}$}
        \STATE $\eta \gets  1-\left (\frac{T}{T_{max}} \right)^ \gamma$\\
        \STATE $\mathcal{B}=\left\{\left(\boldsymbol{x}_{i}, y_{i}\right)\right\}_{i=1}^{batch\_size } \gets $ sample a batch from $\mathcal{D}$;
        \STATE $\boldsymbol{z^{(1)}}, \boldsymbol{z^{(2)}} \gets \textit{proj}_{1}(\phi(\boldsymbol{x^{(1)}})), \textit{proj}_{1}(\phi(\boldsymbol{x^{(2)}}))$ 
        \STATE $\boldsymbol{p^{(1)}}, \boldsymbol{p^{(2)}} \gets \textit{f}(\phi(\boldsymbol{x^{(1)}})), \textit{f}(\phi(\boldsymbol{x^{(2)}}))$
        \STATE $\boldsymbol{h^{(1)}}, \boldsymbol{h^{(2)}} \gets \textit{proj}_{2}(\boldsymbol{z^{(1)}}), \textit{proj}_{2}(\boldsymbol{z^{(2)}})$
        \STATE Calculate class means $\boldsymbol{\mu}^{(1)}, \boldsymbol{\mu}^{(2)}$
        \STATE Based on Eq.\ref{hycon_loss}, calculate $\mathcal{L}_{con}$
        \STATE Based on Eq. \ref{lpu}, calculate  $\mathcal{L}_{p}(\tilde{\boldsymbol{\mu}})$
        \STATE Based on Eq. \ref{lpw}, calculate  $\mathcal{L}_{p}(\boldsymbol{\omega })$
        \STATE Based on Eq. \ref{cls}, calculate  $\mathcal{L}_{cls}^{(1)}, \mathcal{L}_{cls}^{(2)}$\\
        
        // \emph{Calculate the total loss}
        \STATE
        $
\mathcal{L}_{all} = \mathcal{L}_{cls}^{(1)} + \mathcal{L}_{cls}^{(2)} + \alpha (\mathcal{L}_{con} + \mathcal{L}_{p}(\tilde{\boldsymbol{\mu}}))$

        \STATE Update model parameters
        \ENDFOR
    \end{algorithmic}
\end{algorithm}

\section{Experiments} 
In our experiments, we focus on a classic task in imbalanced learning, \textit{i.e.}, long-tailed classification. Next, we describe the experiments in this paper in detail.
\subsection{Datasets}
In the following experiments, we use two balanced datasets (CIFAR10 and CIFAR100) and four long-tailed datasets (CIFAR10-LT, CIFAR100-LT, ImageNet-LT and iNaturalist2018) to verify the effectiveness of the proposed method.

\subsubsection{CIFAR10/100 and CIFAR10/100-LT} Both CIFAR10 and CIFAR100 have 60,000 images, where 50,000 are used for training and 10,000 for testing, corresponding to 10 and 100 categories respectively\cite{krizhevsky2009learning}. CIFAR10-LT and CIFAR100-LT are subsets of CIFAR10 and CIFAR100 respectively. 
For a fair comparison, we construct CIFAR10/100-LT using the same Settings as in \cite{cao2019learning,cui2019class,zhu2022balanced}.
Specifically, they are created by downsampling the samples of each category to obey an exponential decay with an imbalanced ratio $\beta$. Here $\beta=N_{max}/N_{min}$, which
reflects the degree of the imbalance. We evaluated the proposed method on challenging dataset settings ($\beta=10,50,100,200$).

\subsubsection{ImageNet-LT} ImageNet-LT is a long-tailed version of the original ImageNet-2012\cite{deng2009imagenet}, which samples a subset according to a Pareto distribution. In general, it has 115.8K images from 1000 categories, with a maximum of 1280 images per category and a minimum of 5 images per category.

\subsubsection{iNaturalist2018} iNaturalist 2018 is a large-scale classification dataset with extremely imbalanced label distribution collected from the real world\cite{van2018inaturalist}. It consists of 437.5K images from 8142 categories.

All the data sets involved in the experiment are summarised in Table \ref{dataset}.

% Table generated by Excel2LaTeX from sheet 'Sheet3'
\begin{table*}
  \centering
  \caption{Detailed statistical information of the datasets used for the experiment}
    \begin{tabular}{l|ccccccc}
    \toprule
    Dataset & Training Samples & Test Samples & Classes & Max Class Size & Min Class Size & Imba. Factor $\beta$ & Annotation Types \\
    \midrule
    CIFAR10\cite{krizhevsky2009learning} & 50000 & 10000 & 10    & 5000  & 5000 & 1 & Classification \\
    CIFAR100\cite{krizhevsky2009learning} & 50000 & 10000 & 100   & 500   & 500  & 1 & Classification \\
    CIFAR10-LT\cite{cui2019class} & 11203-50000 & 10000 & 10    & 5000  & 25-500 & 10-200 & Classification \\
    CIFAR100-LT\cite{cui2019class} & 11203-50000 & 10000 & 100   & 500   & 2-50  & 10-200 & Classification \\
    ImageNet-LT\cite{deng2009imagenet} & 115846 & 50000 & 1000  & 1280  & 5     & 256   & Classification  \\
    iNaturalist2018\cite{van2018inaturalist} & 437513 & 24426 & 8142  & 1000  & 2     & 500   & Classification \\
    \bottomrule
    \end{tabular}%
  \label{dataset}%
\end{table*}%

% Table generated by Excel2LaTeX from sheet 'Sheet2'
\begin{table*}
  \centering
      \caption{Comparative results of top-1 accuracy (\%) on CIFAR-10/100-LT, with the best results shown in bold.}
    \begin{tabular}{l|rrrr|rrrr}
    \toprule
    \multirow{2}[4]{*}{Method} & \multicolumn{4}{c|}{CIFAR10-LT} & \multicolumn{4}{c}{CIFAR100-LT} \\
\cmidrule{2-9}          & 200   & 100   & 50    & 10    & 200   & 100   & 50    & 10 \\
    \midrule
    CE Loss & 65.68 & 70.70  & 74.81 & 86.40  & 34.84 & 38.43 & 43.90  & 55.80 \\
    \midrule
    LDAM-DRW~\cite{cao2019learning}   & 73.52 & 77.03 & 81.03 & - & 34.84 & 38.43 & 43.90 & - \\
    BBN~\cite{zhou2020bbn}   & 73.47 & 79.82 & 81.18 & 88.32 & 37.21 & 42.56 & 47.02 & 59.12 \\
    LWS+mixup~\cite{zhong2021improving} & -     & 76.30  & 82.60  & 89.60  & -     & 47.00    & 52.30  & 63.20 \\
    MisLAS~\cite{zhong2021improving} & 77.31 & 82.06 & 85.16 & -     & 42.33 & 47.50  & 52.62 & - \\
    RISDA~\cite{chen2022imagine} & 74.00    & 79.89 & 84.24 & 89.36 & 44.76 & 50.16 & 53.84 & 62.38 \\
    BCL~\cite{zhu2022balanced}   & -     & 84.32 & 87.24 & 91.12 & -     & 51.93 & 56.59 & 64.87 \\
    NC Classifier~\cite{yang2022inducing} & 71.90  & 76.50  & 81.00    & -     & 40.90  & 45.30  & 50.40  & - \\
    SBCL~\cite{hou2023subclass} & -  & -  & -    & -     &-  & 44.90  & 48.70  & 57.90 \\
    GLMC~\cite{du2023global}  & -     & 87.75 & 90.18 & 94.04 & -     & 55.88 & 61.08 & 70.74 \\
    CSA+mixup~\cite{shi2023re} & -     & 82.50  & 86.00    & 90.80  & -     & 46.60  & 51.90  & 62.60 \\
    GCL+CR~\cite{ma2023curvature} & 79.90  & 83.50  & 86.80  & -     & 45.60  & 49.80  & 55.10  & - \\
    PC~\cite{sharma2023learning}    & -     & -     & -     & -     & -     & 53.40  & 57.75 & 69.12 \\
    NC+DRW+cRT~\cite{liu2023inducing} & -     & 82.60  & -     & 90.20  & -     & 48.70  & -     & 63.60 \\
    \midrule
    AllNC & \textbf{85.46} & \textbf{88.77} & \textbf{92.12} & \textbf{95.22} & \textbf{51.29} & \textbf{56.63} & \textbf{62.94} & \textbf{73.48} \\
    AllNC++ & \textbf{86.08} & \textbf{90.12} & \textbf{92.43} & \textbf{95.56} & \textbf{51.50}  & \textbf{57.01} & \textbf{62.42} & \textbf{73.60} \\
    \bottomrule
    \end{tabular}%
  \label{result-1}%
\end{table*}%

\subsection{Training Details}

We performed imbalanced classification experiments on CIFAR10-LT and CIFAR100-LT, and balanced classification experiments on CIFAR10 and CIFAR100, with ResNet-32 as the network architecture. All models were trained with the same training setup. Specifically, we train for 200 epochs, using the SGD optimizer with a momentum of 0.9, setting the initial learning rate to 0.01, weight decay to 5e-3, and batch size to 64. Following \cite{cui2021parametric}, we choose Random Augmentation for our augmentation policy.

We also performed imbalanced classification experiments on ImageNet-LT. We use ResNeXt-50-32x4d as the backbone, with an initial learning rate set to 0.1, weight decay to 2e-4 and batch size set to 128. In addition, we also incorporate a large-scale imbalanced dataset, iNaturalist 2018. It is inherently long-tailed and has an extremely unbalanced distribution. We use ResNet50 as the backbone and keep the rest of the settings consistent with ImageNet-LT.

In all experiments, we used Generalized Re-weighting~\cite{zhang2021distribution} as the rebalancing loss, and other training protocols and augmentation strategies mainly followed \cite{zhu2022balanced}. 

\textbf{Evaluation protocol.} For each imbalanced dataset, we train them on a long-tailed training set and evaluate them on a balanced validation/test set. Following \cite{zhu2022balanced}, we further report the accuracies on three splits of classes, including Many-shot classes (training samples $>$ 100), Medium-shot (training samples 20 $\sim $ 100), and Few-shot (training samples $\le $ 20), to comprehensively evaluate our models.

\subsection{Comparison Results}
\paragraph{Compared Methods.} To validate the effectiveness of our AllNC, we selected various recently published representative methods for experimental comparison, including baseline methods (CE Loss), decoupled methods (LDAM-DRW~\cite{cao2019learning}, BBN~\cite{zhou2020bbn}, MisLAS~\cite{zhong2021improving}, GCL/MBJ+CR~\cite{ma2023curvature} and PC~\cite{sharma2023learning}), contrastive strategies (PaCo~\cite{cui2021parametric}, BCL~\cite{zhu2022balanced}, SBCL~\cite{hou2023subclass} and GLMC~\cite{du2023global}), augmentation strategies (LWS+mixup~\cite{zhong2021improving}, RISDA~\cite{chen2022imagine} and CSA+mixup~\cite{shi2023re}), and NC recovery-based methods (NC classifier~\cite{yang2022inducing} and NC+DRW+cRT~\cite{liu2023inducing}). 
Additionally,  a weight rebalancing strategy \cite{alshammari2022long} is added to AllNC, which prompts the learned classifier vectors with approximately equal paradigms, termed as AllNC++.

\subsubsection{Experimental Results on Imbalanced Classification}
\paragraph{CIFAR-10/100-LT} The results on the CIFAR-10/100-LT dataset are summarized in Tab. \ref{result-1}, where we set multiple imbalance ratios: 200, 100, 50 and 10. Across all ratios, our method consistently outperforms the compared methods by a significant margin. Specifically, when $\beta=200$, the top-1 classification accuracies of CIFAR-10-LT and CIFAR-100-LT reach 86.08\% and 51.50\%, which improve the top-1 classification accuracy by \textbf{6.18\%} and \textbf{5.90\%} compared to the suboptimal method GCL+CR, respectively. 
Additionally, when $\beta=10$, the top-1 classification accuracy of AllNC is improved by \textbf{5.36\%} and \textbf{10.00\%}, respectively, compared to the latest method NC+DRW+cRT, which recovers NC from one side. This demonstrates that fully restoring NC from multiple aspects can significantly improve long-tailed classification performance.

% Table generated by Excel2LaTeX from sheet 'Sheet2'
\begin{table}[t]
  \centering
  \caption{Comparative results of top-1 accuracy (\%) on ImageNet-LT, with the best results shown in bold.}
    \begin{tabular}{l|rrrr}
    \toprule
    \multirow{2}[4]{*}{Method} & \multicolumn{4}{c}{ImageNet-LT} \\
\cmidrule{2-5}          & Many  & Medium & Few   & All \\
    \midrule
    CE Loss   & 64.0    & 33.8  & 5.8   & 41.6 \\
    LDAM~\cite{cao2019learning}  & 60.4  & 46.9  & 30.7  & 49.8 \\
    LWS+mixup~\cite{zhong2021improving} & 66.1  & 52.2  & 34.5  & 54.6 \\
    MisLAS~\cite{zhong2021improving} & 61.7  & 51.3  & 35.8  & 52.7 \\
    PaCo~\cite{cui2021parametric} & 64.4  & 54.7  & 33.7  & 56.0 \\
    SBCL~\cite{hou2023subclass} & 63.8  & 51.3  & 31.2  & 53.4 \\
    GLMC~\cite{du2023global} & \textbf{70.1}  & 52.4  & 30.4  & 56.3 \\
    CSA~\cite{shi2023re} & 63.6  & 47.0    & 23.8  & 49.7 \\
    MBJ+CR~\cite{ma2023curvature} & 62.8  & 49.2  & 40.4  & 53.4 \\
    PC~\cite{sharma2023learning} & 63.5  & 50.8  & \textbf{42.7}  & 54.9 \\
    NC+DRW+cRT~\cite{liu2023inducing} & 65.6  & 51.2  & 35.4  & 54.2 \\
    \midrule
    AllNC &   68.0    &  54.5     &   32.6    & \textbf{56.5} \\
    AllNC++ &   67.9    & \textbf{54.9}     &  33.7     & \textbf{56.9} \\
    \bottomrule
    \end{tabular}%
  \label{result-2}%
\end{table}%

\paragraph{ImageNet-LT} As shown in Tab. \ref{result-2}, our method and the compared methods demonstrate Many, Medium, Few and All top-1 classification accuracies on the ImageNet-LT dataset. Our algorithm achieves significant performance improvements in the \textit{All}, which can be attributed to the focus of AllNC on comprehensively restoring NC, aiming to balance classification accuracy across all classes rather than individual majority or minority classes. In particular, compared to the latest algorithm for NC recovery, NC-DRW-cRT, our method achieves a \textbf{2.7\%} improvement. It is worth noting that almost all methods have struggled to attain consistent SOTA performance across all criteria.  For example, PC achieves SOTA on ”Few” but suffers a significant drop in ”Many,” scoring only 63.5\%, surprisingly lower than CE’s 64.0\%, resulting in inferior overall performance.

% Table generated by Excel2LaTeX from sheet 'Sheet2'
\begin{table}[t]
  \centering
  \caption{Comparative results of top-1 accuracy (\%) on iNaturalist 2018, with the best results shown in bold.}
    \begin{tabular}{l|cccc}
    \toprule
    \multirow{2}[4]{*}{Method} & \multicolumn{4}{c}{iNaturalist 2018} \\
\cmidrule{2-5}          & Many  & Medium & Few   & All \\
    \midrule
    CE Loss & 67.2  & 63    & 56.2  & 61.7 \\
    LDAM~\cite{cao2019learning}  & -     & -     & -     & 64.6 \\
    BBN+CR~\cite{ma2023curvature} & 50.6  & 71.5  & 66.8  & 67.6 \\
    LWS+mixup~\cite{zhong2021improving} & 72.8  & 71.6  & 69.8  & 70.9 \\
    RISDA~\cite{chen2022imagine} & -     & -     & -     & 69.2 \\
    PC~\cite{sharma2023learning}    & 71.6  & 70.6  & 70.2  & 70.6 \\
    SBCL~\cite{hou2023subclass}  & \textbf{73.3}  & 71.9  & 68.6  & 70.8 \\
    GLMC~\cite{du2023global}  & -     & -     & -     & - \\
    MBJ+CR~\cite{ma2023curvature} & 73.1  & 70.3  & 66.0    & 70.8 \\
    NC+DRW+cRT~\cite{liu2023inducing} & -     & -     & -     & - \\
    \midrule
    AllNC & 72.6  & \textbf{72.0}    & \textbf{70.8}  & \textbf{71.5} \\
    AllNC++ & 72.3  & \textbf{72.2}    & \textbf{71.6}  & \textbf{71.9} \\
    \bottomrule
    \end{tabular}%
  \label{ina-top1}%
\end{table}%

\paragraph{iNaturalist2018} We train our method on the iNaturalist2018 dataset and the experimental results are summarized in Tab 
\ref{ina-top1}.  Compared to other methods, our proposed AllNC obtains the highest top-1 classification accuracy, which is 0.6\% higher than the sub-optimal result and 9.8\% higher than the baseline method.
In addition, AllNC significantly improves the classification accuracy of minority classes, indicating that our model gives more attention to minority classes to alleviate the \textit{complete minority collapses} problem. Meanwhile, we can observe that AllNC has smaller differences in classification accuracies in the Many, Medium and Few groups, which further suggests that our method obtains more balanced classification accuracies by fully recovering NC.

\begin{table}[!t]
  \centering
  \caption{Top-1 accuracy (\%) on the complete CIFAR-10/100 dataset.}
    \begin{tabular}{l|r|r|r}
    \toprule
    Method & Augmentation & CIFAR-10 & CIFAR-100 \\
    \midrule
    vanilla & Simple Augment & 94.85 & 75.28 \\
    vanilla & MixUp & 95.95 & 77.99 \\
    vanilla & CutMix & 95.41 & 78.03 \\
    GLMC~\cite{du2023global}  & MixUp + CutMix & 97.23 & 83.05 \\
    \midrule
    Supcon~\cite{khosla2020supervised} & RandAugment & 96.00    & 76.50 \\
    PaCo~\cite{cui2021parametric}  & RandAugment & -     & 79.10 \\
    \midrule
    AllNC & RandAugment & \textbf{97.57} & \textbf{83.78} \\
    \bottomrule
    \end{tabular}%
  \label{balance}%
\end{table}%
\subsubsection{Experimental Results on Balanced Classification}
To validate the efficacy of AllNC on balanced datasets,  we perform experiments on CIFAR10/100 and compare it with the latest relevant methods (Mixup~\cite{zhang2018mixup}, CutMix~\cite{yun2019cutmix}, SupCon~\cite{khosla2020supervised}, PaCo~\cite{cui2021parametric}, GLMC~\cite{du2023global}). As detailed in Tab. \ref{balance}, AllNC obtains consistent improvements on different datasets. In particular, on the CIFAR100 dataset, AllNC achieves a classification accuracy of  83.78\%, which is \textbf{4.68\%} better than PaCo using the same augmentation approach, significantly improving the model performance. Additionally, we follow the vanilla methods using \textit{ResNet32} as the backbone, while SupCon, PaCo and GLMC all employ \textit{ResNet50}.

\subsection{Systematic analysis of the proposed method}
\label{reproduction}

\paragraph{Escaping the Complete Minority Collapses}
To verify the effectiveness of our proposed method, we compute the Stds of the corresponding cosine values of the class means and classifier vectors obtained using AllNC  (As shown by green lines in Fig. \ref{avg_cos}). It can be found that the corresponding Stds are significantly lower and close to the balanced ones when using our method. This suggests that AllNC is able to alleviate the \textit{complete minority collapses} problem and obtains models that are closer to NC.

\begin{figure}[!t]
  \centering
   \includegraphics[width=0.9\linewidth]{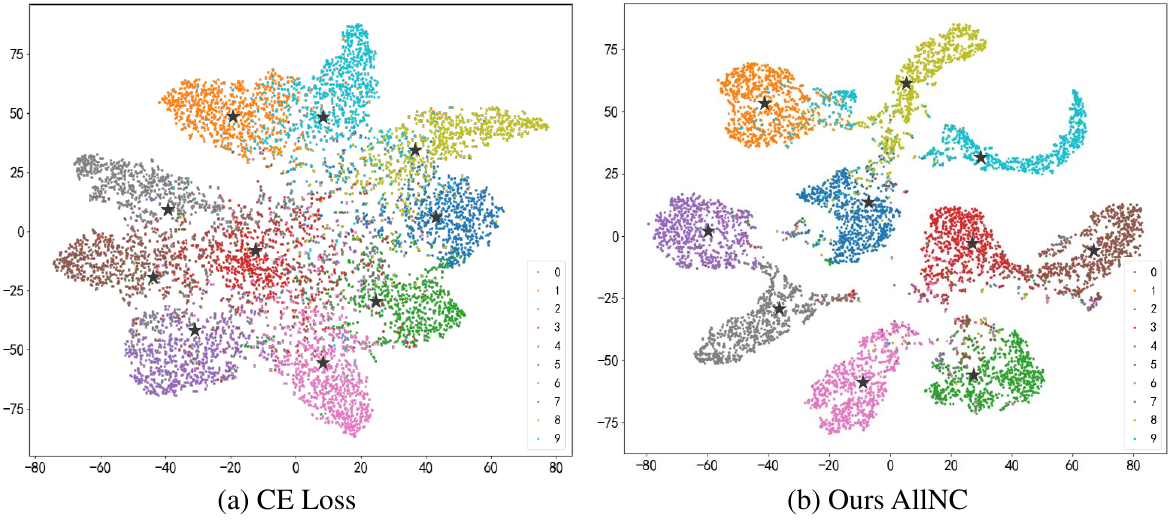}
   \caption{Visualisation results of the last layer of individual activations obtained using different methods.}
   \label{tSNE}
\end{figure}

\paragraph{Reproduction of NC1}
In Fig. \ref{tSNE}, we illustrate the features obtained using CE loss and AllNC separately. Our method achieves superior intra-class tightness for individual activations and sharp categorization boundaries for clusters when compared to CE loss. This suggests that AllNC effectively promotes the collapse of individual activations towards their corresponding intra-class means ($\star$) in an imbalanced context, thereby satisfying the NC1 characteristic.

\begin{figure}
  \centering
   \includegraphics[width=0.9\linewidth]{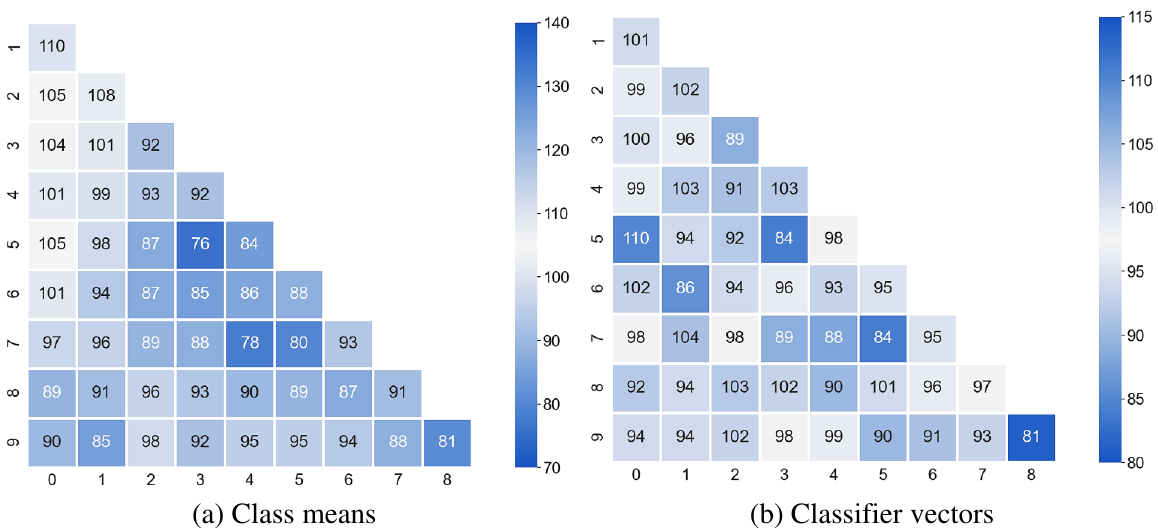}
   \caption{Inter-Class Pairwise Angles (ICPAs) calculated using AllNC.}  
\label{AllNC_angle} 
\end{figure}

\paragraph{Reproduction of NC2 and NC3}
We compute the ICPAs corresponding to the individual activation and classifier vectors obtained using AllNC on CIFAR10-LT. In Fig. \ref{AllNC_angle}, the color intensity of all grids is noticeably lighter compared to (b) and (d) in Fig. \ref{nc2_fig}, indicating that the ICPAs computed using our method are close to the optimal value of $96.4^\circ$.  This suggests that AllNC effectively mitigates \textit{complete minority collapses}, leading to the convergence of class means and classifier vectors towards a simplex ETF structure that satisfies NC2 and NC3. Furthermore, by combining \textbf{Reproduction of NC1}, our method can approximate comprehensive recovery of NC in imbalanced contexts.

\subsection{Ablation studies}

\begin{table*}[t]
  \centering
    \caption{The top-1 classification accuracies for different parameters $\gamma$ on the CIFAR100-LT dataset with imbalance rate $\beta=200$ and $\beta=100$}
    \begin{tabular}{l|rrrr|rrrr}
    \toprule
    \multirow{3}[6]{*}{$\gamma$} & \multicolumn{8}{c}{CIFAR100\_LT} \\
\cmidrule{2-9}          & \multicolumn{4}{c|}{$\beta$=200} & \multicolumn{4}{c}{$\beta$=100} \\
\cmidrule{2-9}          & Many  & Medium & Few   & All   & Many  & Medium & Few   & All \\
    \midrule
    0.5   & 78.83 & 59.83 & 21.55 & 50.19 & 77.43 & \textbf{59.38} & 25.94 & 55.30 \\
    1.0   & \textbf{81.10}  & 57.87 & 21.17 & 50.16 & \textbf{79.54} & 56.94 & 26.68 & 55.47 \\
    1.5   & 79.73 & 58.97 & 23.90  & 51.17 & 79.26 & 56.18 & 27.58 & 55.37 \\
    2.0   & 80.10  & 57.94 & \textbf{24.70}  & \textbf{51.29} & 78.91 & 56.62 & 28.16 & 55.60 \\
    2.5   & 80.67 & 56.87 & 23.82 & 50.79 & 78.94 & 55.26 & 30.07 & 55.72 \\
    3.0   & 81.00    & 56.90  & 23.59 & 50.81 & 77.66 & 56.18 & 33.39 & \textbf{56.63} \\
    3.5   & 80.10  & 58.43 & 23.86 & 51.11 & 75.20  & 55.38 & \textbf{33.58} & 55.56 \\
    4.0   & 79.96 & \textbf{60.59} & 22.48 & 51.16 & 76.86 & 56.41 & 33.26 & 56.40 \\
    \bottomrule
    \end{tabular}%
  \label{ga}%
\end{table*}%

\subsubsection{The Effect of the Adjustable Parameter $\gamma$}
In GBBN, a component of AllNC, we can use the parameter $\gamma$ to regulate the attention of the model to train features and classifiers. As shown in Fig. ~\ref{framework}(b), a larger $\gamma$ means that the model pays more attention to the training of features. To explore the effect of parameter $\gamma$ on model performance, we use different $\gamma$ to train the model on the CIFAR100-LT dataset with an imbalance rate $\beta=100/200$. 
In Tab. \ref{ga}, the top-1 classification accuracies of the model are detailed for various $\gamma$ settings.  It is evident that the accuracies for the Many, Medium, Few, and All groups are responsive to changes in $\gamma$. When the overall performance of the model is optimal, different rates of imbalance are matched with respective $\gamma$ values, which suggests that decoupled feature training and classifier training should be given different attention depending on the degree of imbalance of the dataset.

\subsubsection{The Effect of different components}
\begin{table}[t]
  \centering
  \caption{Ablation results of AllNC on the CIFAR100-LT ($\beta=100$).}
    \begin{tabular}{l|rrrr}
    \toprule
    \multirow{2}[4]{*}{Ablations} & \multicolumn{4}{c}{CIFAR100-LT} \\
\cmidrule{2-5}          & Many  & Medium & Few   & All \\
    \midrule
    AllNC(ours) & 77.66 & 56.18 & 33.39 & \textbf{56.63} \\
    w/o HyCon & 77.14 & 56.09 & 29.75 & 55.21 \\
    w/o P2P & 79.12 & 55.03 & 30.16 & 55.56 \\
    w/o GBBN & 63.26 & 54.65 & 40.19 & 53.18 \\
    w/o  adjustable $\gamma$ & 77.34 & 59.38 & 25.94 & 55.30 \\
    \bottomrule
    \end{tabular}%
  \label{ablation}%
\end{table}%
We conduct an ablation study on the critical components in the AllNC of CIFAR100-LT ($\beta$ = 100), and the detailed results are shown in Tab.\ref{ablation}. We can see that all the components proposed in this paper can significantly improve the overall performance of the model. Furthermore, we can discover that using HyCon and P2P, the accuracy of the Few-shot is improved by \textbf{3.64\%} and \textbf{3.23\%}, respectively, indicating that they can effectively mitigate the \textit{complete minority collapses} problem. When using GBBN, boosting the performance of the Many group is improved by \textbf{14.40\%}, indicating that using the progressively decoupled can greatly mitigate the impact of rebalancing loss on the majority classes. When using the adjustable $\gamma$, the performance of the Few group improves by \textbf{7.45\%}, indicating that adjusting the training bias of the model can obtain a more optimal decoupling process.

\subsubsection{The Effect of the Hyperparameter $\alpha$}
\begin{figure*}[h]
  \centering
   \includegraphics[width=0.9\linewidth]{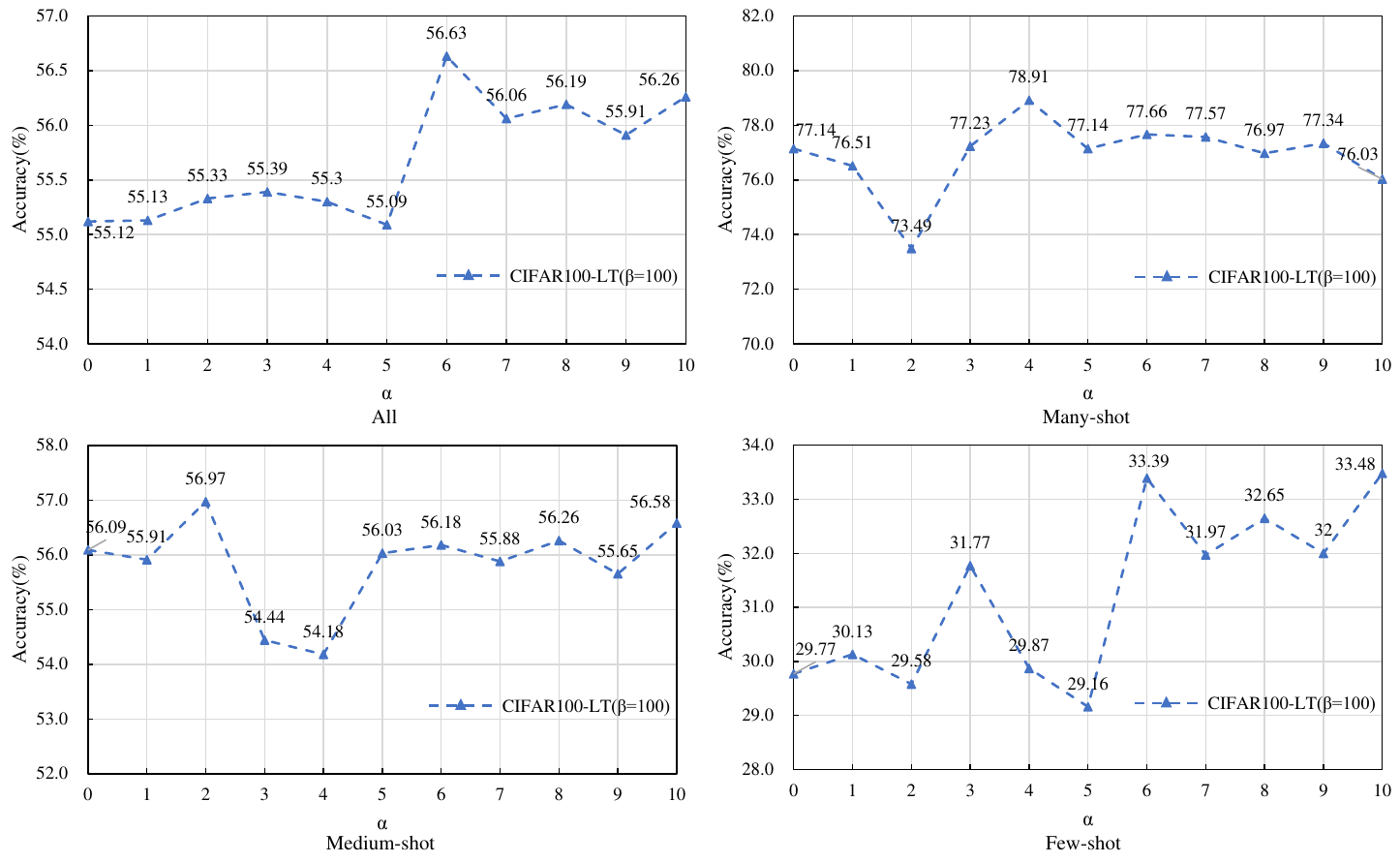}
   \caption{The effect of different hyperparameters $\alpha$ on the performance of different shot groups on the CIFAR100-LT ($\beta=100$) dataset.}
   \label{alpha}
\end{figure*}
In Eq.~\ref{all_loss}, we control the proposed $\mathcal{L}_{con}$ loss and $\mathcal{L}_{p}(\tilde{\boldsymbol{\mu}})$ loss by adjusting the hyperparameter $\alpha$. In this section, we verify the effect of r on the performance of different shot groups on the CIFAR100-LT ($\beta=100$) dataset. As shown in Fig. \ref{alpha}, the overall performance of the model is significantly improved by adjusting $\alpha$. Compared without $\alpha$ (\textit{i.e.}, $\alpha=0$), the overall model performance is improved by 1.51\%. In addition, in the Few-shot group, we can discover that the classification accuracy obtains a significant improvement as $\alpha$ increases. This indicates that the proposed HyCon loss and P2P loss can significantly improve the problem that the majority classes squeeze the classification boundary of the minority classes.

\section{Conclusion}
In this work, we discover that when a model is trained using an imbalanced dataset, NC with optimal separable structure is distorted by \textit{complete minority collapses}, that is, both the feature means and the classifier vectors corresponding to the minority classes are overwhelmed by the majority classes. Therefore, merely restoring NC in any one aspect leads to the model losing sight of the others, making it difficult to recover the inherent \textit{self-duality} property.
To this end, we propose an end-to-end solution strategy, AllNC, to comprehensively recover NC from as various aspects as possible including individual activations, class means and classifier vectors, with a unified framework designed to progressively decouple individual activations and classifier training.
 On multiple balanced and long-tailed benchmark datasets, we thoroughly validate that our AllNC significantly outperforms the compared methods. 
Nevertheless, our work focuses on imbalanced classification tasks, and how to extend to object detection and instance segmentation is also worth exploring.

\section*{Acknowledgments}
The authors would like to thank the support from the National Natural Science Foundation of China (62376126, 62076124, 62106102), the Natural Science Foundation of Jiangsu Province (BK20210292), and the Hong Kong Scholars Program (XJ2023035).



\bibliographystyle{IEEEtran}
\bibliography{main}

\vfill

\end{document}